\documentclass[conference]{IEEEtran}
\IEEEoverridecommandlockouts
\usepackage{cite}
\usepackage{amsmath,amssymb,amsfonts}
\usepackage{algorithm}
\usepackage{algorithmic}
\usepackage{graphicx}
\usepackage{textcomp}
\usepackage{xcolor}

\DeclareMathOperator*{\argmax}{arg\,max}

\def\BibTeX{{\rm B\kern-.05em{\sc i\kern-.025em b}\kern-.08em
    T\kern-.1667em\lower.7ex\hbox{E}\kern-.125emX}}
\begin{document}

\title{Self Regulated Learning Mechanism for Data Efficient Knowledge Distillation\\
}

\author{\IEEEauthorblockN{1\textsuperscript{st} Sourav Mishra}
\IEEEauthorblockA{\textit{Department of Aerospace Engineering} \\
\textit{Indian Institute of Science}\\
Bangalore, India \\
srvmishra832@gmail.com}
\and
\IEEEauthorblockN{2\textsuperscript{nd} Suresh Sundaram}
\IEEEauthorblockA{\textit{Department of Aerospace Engineering} \\
\textit{Indian Institute of Science}\\
Bangalore, India \\
vssuresh@iisc.ac.in}
}

\maketitle

\begin{abstract}
Existing methods for distillation do not efficiently utilize the training data. This work presents a novel approach to perform distillation using only a subset of the training data, making it more data-efficient. For this purpose, the training of the teacher model is modified to include self-regulation wherein a sample in the training set is used for updating model parameters in the backward pass either if it is misclassified or the model is not confident enough in its prediction. This modification restricts the participation of samples, unlike the conventional training method. The number of times a sample participates in the self-regulated training process is a measure of its significance towards the model's knowledge. The significance values are used to weigh the losses incurred on the corresponding samples in the distillation process. This method is named significance-based distillation. Two other methods are proposed for comparison where the student model learns by distillation and incorporating self-regulation as the teacher model, either utilizing the significance information computed during the teacher's training or not. These methods are named hybrid and regulated distillations, respectively. Experiments on benchmark datasets show that the proposed methods achieve similar performance as other state-of-the-art methods for knowledge distillation while utilizing a significantly less number of samples.
\end{abstract}

% \begin{IEEEkeywords}
% component, formatting, style, styling, insert
% \end{IEEEkeywords}

\section{Introduction}
Deep learning models have shown remarkable performance in several fields such as image classification \cite{NIPS2012_c399862d}, object detection \cite{obj}, etc. However, deploying them on edge devices such as mobile phones and an on-board computer is not feasible due to their larger memory footprint. Therefore several methods have been proposed in the literature to address model compression without compromising generalization performance. Based on their assumption about knowledge representation, the methods can be divided into model compression-based methods \cite{Han2015, Han2016, Sandryhaila2013, Denton2014, Wang2016} and knowledge distillation based methods \cite{normal, ZSKD, FSKD, DAFL}.

Model compression-based methods assume that knowledge is contained in the model's weights and reduce the redundancies present in deep models. Neural network pruning was introduced by LeCun in \cite{brainDamage}. Various other methods for compressing neural networks have been proposed in the literature \cite{Han2015, Han2016, Sandryhaila2013, Denton2014, Wang2016}. Model compression-based methods involve iterative pruning and fine-tuning of networks and are often time-consuming processes. 

On the other hand, knowledge distillation based methods assume that the knowledge of a model is captured in its intermediate activations and outputs. Hence, smaller models known as students receive supervision from larger models called teachers and the ground truths. The probabilities assigned by the teacher to the incorrect classes constitute 'dark knowledge,' and it has been shown to improve the generalization ability of student models \cite{normal}. Based on the type of knowledge being transferred,  knowledge distillation methods fall into one of two families - response-based and feature-based. Response based methods \cite{normal, ZSKD, FSKD, DAFL} transfer the knowledge from the teacher to the student by matching the outputs of their last layers, whereas feature-based methods \cite{Romero2015, Heo2019, Zagoruyko2017} supervise the students by matching the activations of the intermediate layers of the teacher and the student models. The original training data is used to perform distillation in \cite{normal}. Methods to construct synthetic samples for distillation are proposed in \cite{FSKD, meta, ZSKD, DAFL}. These methods are based on the conventional way of training models where all the samples participate equally in learning the input-output mapping. Due to their inability to discriminate between samples based on their importance towards learning, these methods are very inefficient in terms of data usage and require large amounts of data.

However, in machine learning literature, it has been shown that metacognitive neural network acheieve better generalization by employing self-regulation to select appropriate training samples for learning from stream-of-training data \cite{meta1, meta2, meta3}. The heuristic strategies help in accounting for the different levels of knowledge present in different samples resulting in improved overall generalization and data-efficiency.

This work address the data-efficiency issue of current state-of-the-art distillation methods by employing self-regulation. The teacher network uses an adaptive threshold to maximize the inter-class posterior probability difference during training. In this process, the samples on which it learns faster get filtered out from further training. The participation of each sample in training is monitored and is used to compute its significance value, which is a measure of its contribution to the teacher model's knowledge. During knowledge transfer, the student model's learning is driven by applying the computed sample significance information (sample significance based distillation), or by self-regulation alone (regulated distillation), or by a combination of both (hybrid distillation). The proposed methods (summarized in Figure 1) are data-efficient as they utilize significantly fewer training samples than other methods for knowledge distillation. The proposed distillation methods are evaluated on three benchmark data sets - MNIST, Fashion-MNIST, and CIFAR10. The results establish the data efficacy of the proposed distillation methods and their competitive performance with current state-of-the-art results reported in the literature.

The main contributions of the work is summarized below:
\begin{itemize}
    \item For the first time in distillation literature, the data-efficiency issue is addressed.
    \item Self-regulation is proposed as a technique to improve data-efficiency as it accounts for the different levels of knowledge present in different samples.
    \item Three types of data-efficient approaches for knowledge transfer are proposed - sample significance based, regulated and hybrid. In sample significance based distillation, the significance information computed during teacher training is used to guide the student model's learning. In regulated distillation, the student model employs self-regulation to learn from the soft targets produced by the teacher. In the hybrid strategy, both the above mechanisms are combined to guide the student.
    \item The proposed distillation schemes are evaluated on the benchmark datasets - MNIST, Fashion-MNIST, and CIFAR10. The proposed methods achieve similar or slightly better generalization performance than the current state-of-the-art distillation methods while utilizing much less data samples in the process.
\end{itemize}
    
\section{Related Works}

The idea of distillation was proposed in \cite{org} and gained momentum in \cite{normal}. The student model is found to generalize better if supervised by the soft targets obtained at a high temperature from a bigger teacher model instead of the conventional way of training. This provides an easy method for transferring most of the generalization capacity of larger models to smaller models. Research in distillation is motivated by this observation. Apart from model compression, distillation has been successfully used in other applications as well. Recently, distillation has been applied in face recognition \cite{deepFace}, cross-modal hashing \cite{CMHash} and collaborative learning \cite{collaborative}. Several methods have been proposed for distillation and, a comprehensive review is provided in \cite{review}.

\subsection{Knowledge Distillation}
The idea of knowledge distillation was popularised by Hinton in \cite{normal}. It proposed to use the original training data as the transfer set for distillation. In addition to learning from the ground truths, the student also receives supervision from the teacher model in the form of soft targets computed at a high softmax temperature. Subsequently, ~\cite{DAFL, ZSKD, meta} proposed methods for knowledge transfer wherein the transfer set was not available. The softmax space of the teacher network is modeled by Dirichlet distribution in ~\cite{ZSKD}. Synthetic data instances are constructed by inverting samples drawn from this distribution. Using the teacher model as a fixed discriminator to train a generator for constructing synthetic samples is proposed in ~\cite{DAFL}. Constructing synthetic samples from the activation statistics of the teacher model's training is proposed in ~\cite{meta}. The process of distillation is the same as in ~\cite{normal}.

The feature extraction process in students is supervised by the activations of the intermediate layers of a teacher model in ~\cite{Romero2015}. In case the output sizes of the layers involved in the transfer process do not match, a learnable convolutional regressor network is used to match the sizes. Transfer of activations of hidden neurons rather than their actual response values is proposed in ~\cite{Heo2019}. It is shown that the generalization ability is better encoded by the decision boundaries formed by the hidden neurons rather than the actual response magnitudes. Knowledge is transferred from the teacher model to the student model by matching different types of attentions in ~\cite{Zagoruyko2017}. These are computed at certain layers for the teacher and student models and are matched by minimizing the $L_p$ norm of their difference. 

A method for self-distillation is proposed in ~\cite{self-kd1} wherein the teacher and the student models are the same. The model from the previous epoch is used as the teacher. The proposed work is different from self-distillation methods as teacher and student models are different networks of different sizes. Hence, self-distillation methods are not used as baselines for comparison.

\subsection{Self Regulation}
In the conventional deep neural network training and knowledge distillation methods, all the training samples participate equally in capturing the input-output relationship. However, in machine learning literature, it has been shown that regulating the sample participation during training can lead to better generalization \cite{meta3,meta1,meta2}. The conventional method for training disregards the relative importance of each sample in the dataset towards the knowledge of the model. Different samples contain different levels of knowledge. For example, some portions of a book are easier to grasp than others. The reader spends more time on those portions that he finds difficult and less on those portions that he finds easy. Self-regulation emulates this aspect of human learning in neural network training. In this way, the self-regulated learning process is much more efficient in terms of data usage than the conventional training method.

\section{Methods}
In this section, the underlying mathematical and algorithmic details of the proposed data efficient knowledge distillation methods are described.

\begin{figure}[]
    \centering
    \includegraphics[width=0.8\linewidth]{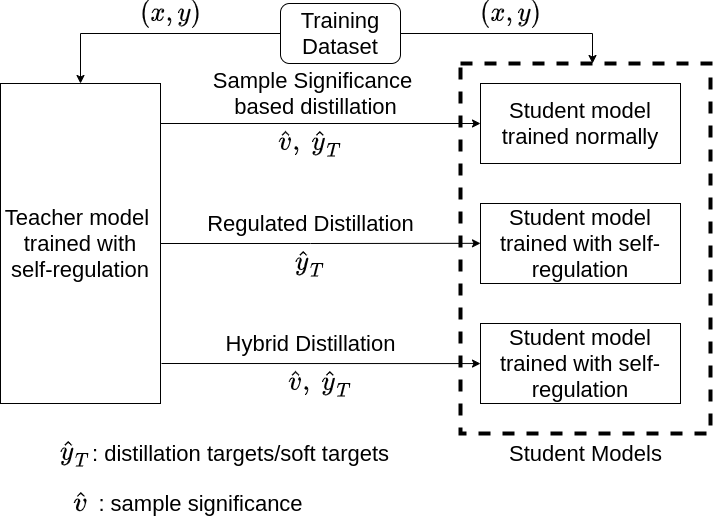}
    \caption{Summary of proposed data-efficient knowledge distillation methods. In sample significance based distillation, the sample significance information computed during teacher training is used to guide the student model. In regulated distillation, the student model learns in the same way as in the conventional distillation method while using self-regulation. In the hybrid method, the student employs self-regulation as well as the sample significance information for learning.}
    \label{fig:my_label}
\end{figure}

\subsection{Self-Regulated Training and Sample Significance Computation}
\subsubsection{Self-Regulated Teacher Training}
Training is made more data-efficient by controlling the participation of samples on which the model is able to learn faster. 

While employing self regulation, the model need not learn on a sample again if it is already too confident on it. So the model is able to distinguish between easy and hard samples based on an epoch dependent threshold and discards the easy samples from the process. In this way, the model learns to focus more on the difficult samples (which contribute more to its knowledge) than easy ones. The self regulation process is explained below.

Given a dataset $\mathbb{D}$ containing labeled samples $(x, y)$ and a model $\mathbb{M}$, the following quantities are monitored for all samples in all epochs $(N)$:

\begin{itemize}
    \item The predicted label, $\hat{y}$: \\$\hat{y} = \argmax \mathbb{M}(x)$
    
    \item The difference between the maximum and the second maximum predicted probabilities, $\delta$: \\$\delta = \max \mathbb{M}(x) - \max \{s | s \in \mathbb{M}(x), \ s \neq \max \mathbb{M}(x)\}$
\end{itemize}

As the model learns to classify properly, the difference $\delta$ gradually increases with the number of epochs $n$. A sample is included used in the backward pass for parameter updates if the predicted class is incorrect or if $\delta$ is less than an epoch dependent adaptive threshold, $\eta$. $\delta$ will increase faster for easy samples compared to difficult samples. The purpose of the epoch dependent threshold function $f(n)$ is to filter out such samples from further training. Since $\delta$ is the difference between the maximum and the second maximum probabilities, it is in the range $[0, 1]$. So the function $f(n): \mathbb{N} \rightarrow [0, 1]$ must be an increasing function of $n$. So $f(n) = 1-\exp(-\alpha n)$ is chosen as threshold predictor, where $\alpha$ is a hyperparameter. It maximizes the difference in the predicted posterior probabilities by allowing samples with a smaller growth rate of $\delta$ to participate more in training. Figure 2 summarizes the method.

\begin{figure}[]
    \centering
    \includegraphics[width=0.8\linewidth]{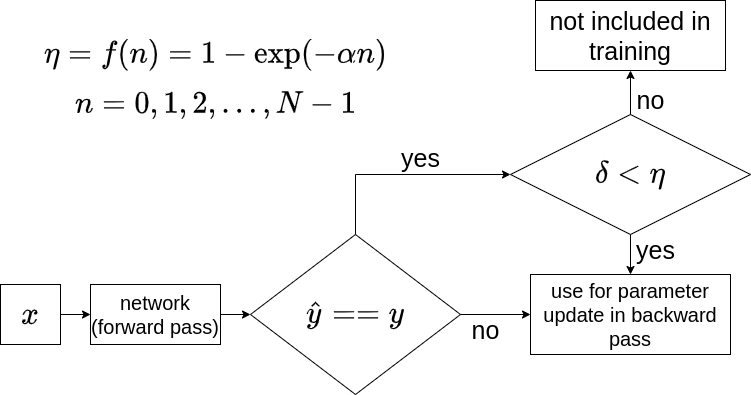}
    \caption{Self Regulated Training Algorithm: A sample is used for parameter update in the backward pass if it is predicted incorrectly by the model, or if the difference between the maximum and the second maximum predicted probabilities ($\delta$) is less than an epoch dependent threshold ($\eta$).}
    \label{fig:my_label}
\end{figure}

\subsubsection{Computation of Sample Significance}
As explained earlier, all the samples in the dataset will not contribute equally to the knowledge of the model. The training process must distinguish samples accordingly to enhance generalization ability of the model. Self-regulation introduced in the previous subsection is a method of doing this. In the conventional training scheme, all the samples present in the dataset $\mathbb{D}$ participate equally. That is, if the model is trained for $N$ epochs, then each sample participates exactly $N$ times in the training. However, with self-regulation, each sample participates $\leq N$ times in the process, with the difficult ones participating more often than the easy ones. In such a scheme, the number of times a sample participates in the training process can be seen as a measure of its contribution to the knowledge of the model. 

The significance of a sample is defined its class-wise min-max normalized participation. It is a number between 0 and 1. Let the significance of a sample be denoted by $\hat{v}$ and the number of times it participates in the training be $v$. Let $S_i$ denote the subset of samples in the dataset $\mathbb{D}$ that belong to the class $i$ out of a total of $C$ classes. That is,

\begin{equation}
    S_i = \{x | (x, y) \in \mathbb{D}, \ \text{and} \ y = i, \ i \in \{0, 1, ..., C-1\}\}
\end{equation}

For a given sample $(x, y)$, the significance $\hat{v}$ is defined by:

\begin{equation}
    \hat{v} = \frac{v - \min_{x \in S_y} v}{\max_{x \in S_y} v - \min_{x \in S_y} v}
\end{equation}

The sample significance information $\hat{v}$ is computed after the self-regulated training of the teacher model as it requires the sample participation data $v$ that is recorded during the teacher's training. It serves as a measure of the importance of the sample to the teacher's knowledge and is used during distillation to transfer the different levels of knowledge present in the different samples. The self regulated teacher training and the sample significance computation processes are described in algorithms 1 and 2, respectively.

\begin{algorithm}
\textbf{Input} Teacher network $T$, with parameters $\theta_T$ (without output softmax), dataset $\mathbb{D} = \{(x_i, y_i)\}_{i=1}^t$, epochs $N$, parameter $\alpha$ for self-regulation\\
\textbf{Output} Parameters of trained Teacher network $\theta_T$, array of sample participations $\mathbf{v}$, \ size of $\mathbf{v}$ is same as $|\mathbb{D}| = t$\\
\begin{algorithmic}[1]
\STATE Initialize participations $\mathbf{v}$
\FOR{$k$ in range($t$)}
\STATE $\mathbf{v}[k] = 0$
\ENDFOR
\STATE Train teacher and record participations
\FOR{$n$ in range($N$):}
\STATE $\eta = 1 - \exp{(\alpha n)}$
\FOR{$j, \ (x, y) \in \text{enumerate(}\mathbb{D})$:}
\STATE $z_T = T(x)$
\STATE $y_T = softmax(z_T)$
% \STATE $z_S = S(x)$
% \STATE $y_S = softmax(z_S/T_0)$
\STATE $\hat{y} = \argmax y_T$
\STATE compute $\delta$ from $y_T$ 
% \STATE $\lambda = y^Ty_T$
\IF{$\hat{y} \neq y$ or ($\hat{y} == y$ and $\delta < \eta$):}
% \STATE compute $y_C$ from equations (4), (5)
\STATE $L = L_{CE}(y, y_T)$
\STATE update teacher's parameters: $\theta_T' = \theta_T - \nabla_{\theta_T} L$
\STATE $\theta_T = \theta_T'$
\STATE $\mathbf{v}[j] = \mathbf{v}[j] + 1$
% \ELSIF{$\hat{y} == y$ and $\delta < \eta$:}
% \STATE compute $y_C$ from equations (4), (5)
% \STATE compute $L$ from equation (6)
% \STATE update student's parameters: $\theta_S' = \theta_S - \nabla_{\theta_S} L$
% \STATE $\theta_S = \theta_S'$
\ELSE
\STATE continue
\ENDIF
\ENDFOR
\ENDFOR
\RETURN $\theta_T, \mathbf{v}$
\end{algorithmic}
\caption{Teacher model training with self-regulation}
\end{algorithm}

\begin{algorithm}
\textbf{Input} Sample Participation Statistics recorded during teacher training $\mathbf{v}$, The Dataset $\mathbb{D} = \{(x_i, y_i)\}_{i=1}^t$ for which the participation is recorded. Dataset $\mathbb{D}$ has a total of $C$ classes labeled as $0, 1, ..., C-1$. \\ 
\textbf{Output} Sample significance vector $\mathbf{\hat{v}}$. The sizes of $\mathbf{\hat{v}}, \mathbf{v}$ and $\mathbb{D}$ are the same.
\begin{algorithmic}[1]
\FOR{$i$ in range($t$)}
\STATE $(x, y) = \mathbb{D}[i]$
\STATE $v = \mathbf{v}[i]$
\STATE compute $\hat{v}$ from equation (2)
\STATE $\mathbf{\hat{v}}[i] = \hat{v}$
\ENDFOR
\RETURN $\mathbf{\hat{v}}$
\end{algorithmic}
\caption{Computation of Sample Significance}
\end{algorithm}

\subsection{Data Efficient Distillation Methods}
\subsubsection{Sample Significance Based Knowledge Distillation}
In conventional knowledge distillation, the soft targets computed from the teacher model at a temperature $\tau$ are used. Given a teacher network $T$ parametrized by $\theta_T$ and a student network $S$ parametrized by $\theta_S$, distillation minimizes the following objective over all samples $(x, y)$ in the transfer set $\mathbb{D}$:

\begin{multline}
    L = \sum_{\substack{(x, y) \in \mathbb{D}}} L_{KD}(S(x, \theta_S, \tau), T(x, \theta_T, \tau)) + \lambda L_{CE}(\hat{y}_S, y)
\end{multline}

where, $L_{KD}$ is the distillation loss which is minimized at a temperature $\tau$. It can be the cross entropy loss for classification or the $L_2$ loss for regression. $L_{CE}$ is the cross entropy loss which is minimized at a temperature of 1. $\hat{y}_S$ is the prediction of the student network on the sample $x$ and $\lambda$ is a hyperparameter to balance the two losses.

In sample significance based distillation, the sample significance information computed above is used to direct the student model's learning along with the soft targets. The loss function thus becomes sample specific and accounts for the different levels of knowledge to be transferred from the teacher model for the different samples in the dataset. The loss incurred on each sample is scaled by its significance computed during teacher training. For sample significance based distillation, the sample significance $\hat{v}$ is also included as a part of the dataset. The loss function is given by:

\begin{multline}
        L_{new} = \sum_{\substack{(x, y, \hat{v}) \in \mathbb{D}}} \hat{v} L_{KD}(S(x, \theta_S, \tau), T(x, \theta_T, \tau)) + \\ \lambda \hat{v} L_{CE}(\hat{y}_S, y)
\end{multline}

In this distillation process, the student receives maximum guidance from the teacher - in the form of soft targets and the sample significance information. As the teacher model is of larger capacity than the student model, it is expected that the samples which were difficult for the teacher will be difficult for the student model as well. So the student must put more focus on such samples during the knowledge transfer process.

\subsubsection{Regulated Knowledge Distillation}
The student model is trained by using the self-regulation strategy proposed in the first subsection but it does not use the sample significance information. In this scheme, the student model is given freedom to discriminate between the samples on its own through self-regulation just like the teacher model. The student model may find a different set of easy and difficult samples compared to the teacher model. The teacher is used to supervise the student through soft targets just like in conventional distillation \cite{normal}.

\subsubsection{Hybrid Knowledge Distillation}
The student model is trained by using the sample significance information as well as by using the proposed self-regulation strategy. In this scheme, two effects are taking place simultaneously. The student model is trying to learn independently through self-regulation and at the same time it receives additional guidance in the form of sample significance information to focus more on the samples that the teacher model found tough during its training. Algorithm 3 shows the implementation of these distillation methods. The distillation methods are summarized in Figure 1.

\begin{algorithm}
\textbf{Input} Pre-trained Teacher network $T$, Student network $S$ with parameters $\theta_S$ (without output softmax), dataset $\mathbb{D} = \{(x_i, y_i)\}_{i=1}^t$, epochs $N$, parameter $\alpha$ for self-regulation, temperature $\tau$ for distillation, hyperparameter $\lambda$, distillation mode - significance, regulated, hybrid, sample significance information (in case of significance based distillation) $\mathbf{\hat{v}}$\\
\textbf{Output} Parameters of trained Student network $\theta_S$\\
\begin{algorithmic}[1]
\FOR{$n$ in range($N$):}
\STATE $\eta = 1 - \exp{(\alpha n)}$
\FOR{$j, \ (x, y) \in \text{enumerate(}\mathbb{D})$:}
\STATE $z_T, \ z_S = T(x), \ S(x)$
\STATE $y_T, \ y_S = softmax(z_T/\tau), \ softmax(z_S/\tau)$
\STATE $y_S' = softmax(z_S)$
\STATE $\hat{y} = \argmax y_S$
\STATE compute $\delta$ from $y_S$ 
\STATE $\hat{v} = \mathbf{\hat{v}}[j]$
\IF{mode == \textbf{regulated}:}
\IF{$\hat{y} \neq y$ or ($\hat{y} == y$ and $\delta < \eta$):}
\STATE $L = L_{KD}(y_T, y_S) + \lambda L_{CE}(y, y_S')$
\STATE update student's parameters: \\$\theta_S' = \theta_S - \nabla_{\theta_S} L$
\STATE $\theta_S = \theta_S'$
\ELSE
\STATE continue
\ENDIF
\ENDIF
\IF{mode == \textbf{significance}:}
\STATE $L = \hat{v}L_{KD}(y_T, y_S) + \lambda \hat{v}L_{CE}(y, y_S')$
\STATE update student's parameters: \\$\theta_S' = \theta_S - \nabla_{\theta_S} L$
\STATE $\theta_S = \theta_S'$
\ENDIF
\IF{mode == \textbf{hybrid}:}
\IF{$\hat{y} \neq y$ or ($\hat{y} == y$ and $\delta < \eta$):}
\STATE $L = \hat{v}L_{KD}(y_T, y_S) + \lambda \hat{v}L_{CE}(y, y_S')$
\STATE update student's parameters: \\$\theta_S' = \theta_S - \nabla_{\theta_S} L$
\STATE $\theta_S = \theta_S'$
\ELSE
\STATE continue
\ENDIF
\ENDIF
\ENDFOR
\ENDFOR
\RETURN $\theta_S$
\end{algorithmic}
\caption{Distillation Algorithms}
\end{algorithm}

\section{Experiments}
This section describes the implementation of the proposed algorithms. The generalization performance and sample efficiency of the proposed methods are evaluated. The MNIST, FashionMNIST, and the CIFAR10 datasets are used for the experiments. The batch size is set to 512, the distillation temperature $\tau$ is set to 20, the normal temperature is set to 1, the hyperparameter $\lambda$ is set to 0.3. The Adam optimizer \cite{adam} is used to train the models. Models are evaluated at the normal temperature. The algorithms are evaluated based on their accuracy on the test set. All implementations are done in pytorch. Two NVIDIA GeForce RTX 2080Ti cards are used for the experiments. The following sections describe the experiments and the results. 

\subsection{Self-Regulated Teacher Training and Data Efficient Distillations}
\subsubsection{Training Teacher Model with Self-Regulation}
First, the teacher models are trained. The results are shown in Table 1. Conventionally training the models is equivalent to setting $\alpha = \infty$ in algorithm 1. It is observed that the self-regulated training performs comparably to the conventional method of training for MNIST and CIFAR10 datasets. In the Fashion-MNIST case, it performs better than the conventional training method for all values of $\alpha$ considered. 

\begin{table}[]
\caption{Test accuracy of teacher model trained using self regulation for different values of $\alpha$.}
\centering
\begin{tabular}{|l|l|l|l|}
\hline
{ $\boldsymbol{\alpha}$} & \textbf{MNIST} & \textbf{FMNIST} & \textbf{CIFAR10} \\ \hline
{ \textbf{0.006}}   & 0.9899                       & 0.9016                        & 0.8273                         \\
{ \textbf{0.008}}   & 0.9911                       & 0.8994                        & 0.8313                         \\
{ \textbf{0.01}}    & 0.9912                       & 0.9006                        & 0.8301                         \\
{ \textbf{0.02}}    & 0.9897                       & 0.9045                        & 0.8281                         \\
{ \textbf{0.04}}    & 0.9902                       & 0.9042                        & 0.8310                         \\
{ \textbf{0.08}}    & 0.9903                       & 0.9018                        & 0.8272                         \\ 
\begin{tabular}[c]{@{}l@{}}$\boldsymbol{\infty}$ \\ \textbf{(normal)}\end{tabular}           & 0.9914                       & 0.8992                        & 0.8325                         \\\hline
\end{tabular}
\label{table:Table1}
\end{table}

% \begin{table}[]
% \caption{Comparison of our distillation methods with other methods available in the literature}
% \centering
% \begin{tabular}{|l|l|l|l|}
% \hline
% \textbf{Method}             & \textbf{MNIST} & \textbf{FMNIST} & \textbf{CIFAR10} \\ \hline
% \begin{tabular}[c]{@{}l@{}}Conventional \cite{normal}\end{tabular}              & 0.9925         & 0.8966          & 0.8008           \\
% Few Shot KD \cite{FSKD}     & 0.8670         & 0.7250          & N/A              \\
% Meta Data \cite{meta}      & 0.9247         & N/A             & N/A              \\
% Data Free KD \cite{DAFL}       & 0.9820         & N/A             & N/A              \\
% Zero Shot KD \cite{ZSKD}     & 0.9877         & 0.7962          & 0.6956           \\ \hline
% \multicolumn{4}{|l|}{\textbf{Ours}}                                               \\ \hline
% Significance based           & 0.9870         & 0.8737          & 0.7079           \\
% Regulated          & 0.9859         & 0.8892          & 0.7234           \\
% Hybrid & 0.9804         & 0.8642          & 0.7266           \\ \hline
% \end{tabular}
% \label{table:Table2}
% \end{table}

\subsubsection{Distillation Results}
The details of the datasets, the teacher and student model sizes and some of the hyperparameters are shown in table 2. The number of epochs is kept the same across teacher training and distillation processes.

\begin{table}[]
\centering
\caption{Details of datasets and other hyperparameters}
\begin{tabular}{|c|c|c|c|}
\hline
                           & \textbf{MNIST}                                                            & \textbf{FMNIST}                                                           & \textbf{CIFAR10}                                                           \\ \hline
\textbf{Training Set Size} & 60000                                                                     & 60000                                                                     & 50000                                                                      \\
\textbf{Testing Set Size}  & 10000                                                                     & 10000                                                                     & 10000                                                                      \\
\textbf{Sample Details}    & \begin{tabular}[c]{@{}c@{}}28x28 \\ grayscale\end{tabular}                & \begin{tabular}[c]{@{}c@{}}28x28 \\ grayscale\end{tabular}                & \begin{tabular}[c]{@{}c@{}}32x32\\ RGB images\end{tabular}                 \\
\textbf{Teacher model}     & \begin{tabular}[c]{@{}c@{}}LeNet-5\cite{LeCun1998}\\ ($\sim$62K params)\end{tabular}      & \begin{tabular}[c]{@{}c@{}}LeNet-5\cite{LeCun1998}\\ ($\sim$62K params)\end{tabular}      & \begin{tabular}[c]{@{}c@{}}AlexNet\cite{NIPS2012_c399862d}\\ ($\sim$1.66M params)\end{tabular}     \\
\textbf{Student model}     & \begin{tabular}[c]{@{}c@{}}LeNet-5 Half\\ ($\sim$36K params)\end{tabular} & \begin{tabular}[c]{@{}c@{}}LeNet-5 Half\\ ($\sim$36K params)\end{tabular} & \begin{tabular}[c]{@{}c@{}}AlexNet Half\\ ($\sim$0.4M params)\end{tabular} \\
\textbf{Epochs}            & 200                                                                       & 200                                                                       & 1000                                                                       \\
\textbf{Teacher LR}        & 0.001                                                                     & 0.001                                                                     & 0.001                                                                      \\
\textbf{Distillation LR}   & 0.01                                                                      & 0.01                                                                      & 0.001                                                                      \\
$\boldsymbol{\alpha}$           & 0.02                                                                      & 0.04                                                                      & 0.04                                                                       \\ \hline
\end{tabular}
\end{table}

Tables 3-5 compare the test accuracies of the proposed methods against those of other response-based knowledge distillation methods available in the literature for the MNIST, Fashion-MNIST, and CIFAR10 datasets.

\begin{table}[]
\centering
\caption{Results on the MNIST Dataset}
\begin{tabular}{|l|l|}
\hline
\textbf{Method}                             & \textbf{Test Accuracy} \\ \hline
Conventional \cite{normal} & 0.9925         \\
Few Shot KD \cite{FSKD}    & 0.8670         \\
Meta Data \cite{meta}      & 0.9247         \\
Data Free KD \cite{DAFL}   & 0.9820         \\
Zero Shot KD \cite{ZSKD}   & 0.9877         \\ \hline
\textbf{Ours}                               &                \\ \hline
Significance based                          & 0.9870         \\
Regulated                                   & 0.9859         \\
Hybrid                                      & 0.9804         \\ \hline
\end{tabular}
\end{table}

% \subsubsection{Distillation Results on FashionMNIST}
% It is a dataset of 10 fashion items. Each sample is a 28$\times$ 28 grayscale image. The training set has 60000 labeled samples and the test set has 10000 labeled samples. Lenet-5 model \cite{LeCun1998} is used as the teacher network and Lenet-5 half model as the student network. The models are trained for 200 epochs. The teacher model is trained with a learning rate of 0.001 and $\alpha = 0.04$ at the normal temperature. A learning rate of 0.01 is used for distillation. The same value of $\alpha$ is used in the 'regulated' and 'hybrid' distillation processes. Table 3 compares the results of the proposed methods against other response based knowledge distillation methods available in the literature.

\begin{table}[]
\centering
\caption{Results on the FashionMNIST Dataset}
\begin{tabular}{|l|l|}
\hline
\textbf{Method}                             & \textbf{Test Accuracy} \\ \hline
Conventional \cite{normal} & 0.8966          \\
Few Shot KD \cite{FSKD}    & 0.7250          \\
Zero Shot KD \cite{ZSKD}   & 0.7962          \\ \hline
\textbf{Ours}                               &                 \\ \hline
Significance based                          & 0.8737          \\
Regulated                                   & 0.8892          \\
Hybrid                                      & 0.8642          \\ \hline
\end{tabular}
\end{table}

% \subsubsection{Distillation Results on CIFAR10}
% It is a dataset of 10 items as classes and each class contains 6000 samples. Each sample is a 32 $\times$ 32 colour image. The training set contains 50000 labeled sapmles and the test set contains 10000 labeled samples. Alexnet \cite{NIPS2012_c399862d} model is used as the teacher network and Alexnet half model as the student network. The models are trained for 1000 epochs. The teacher model is trained with a learning rate of 0.001 and $\alpha = 0.04$ at the normal temperature. A learning rate of 0.001 is used for distillation. The same value of $\alpha$ is used in the 'regulated' and 'hybrid' distillation processes. Table 4 compares the results of the proposed methods against other response based knowledge distillation methods available in the literature.

\begin{table}[]
\centering
\caption{Results on the CIFAR10 Dataset}
\begin{tabular}{|l|l|}
\hline
\textbf{Method}                             & \textbf{Test Accuracy} \\ \hline
Conventional \cite{normal} & 0.8008           \\
Zero Shot KD \cite{ZSKD}   & 0.6956           \\ \hline
\textbf{Ours}                               &                  \\ \hline
Significance based                          & 0.7079           \\
Regulated                                   & 0.7234           \\
Hybrid                                      & 0.7266           \\ \hline
\end{tabular}
\end{table}

On simpler datasets such as MNIST, there isn't much difference in performance across the distillation methods. On more realistic datasets like CIFAR10, hybrid distillation might perform slightly better than others. Regulated and hybrid distillations are expected to perform better on realistic scenarios because the student model is given the freedom to discriminate between the samples from the dataset through self-regulation. In this way, it can learn in a better way. The proposed methods perform better than most of the state-of-the-art methods (Tables 3-5). However, \cite{normal} performs better than the proposed methods because it uses all the samples available in the dataset. The advantage of the proposed methods is that they do not use all the samples. They are highly efficient in terms of data usage, as explained in the next section.

\begin{figure*}[]
    \centering
    \includegraphics[width=0.8\linewidth]{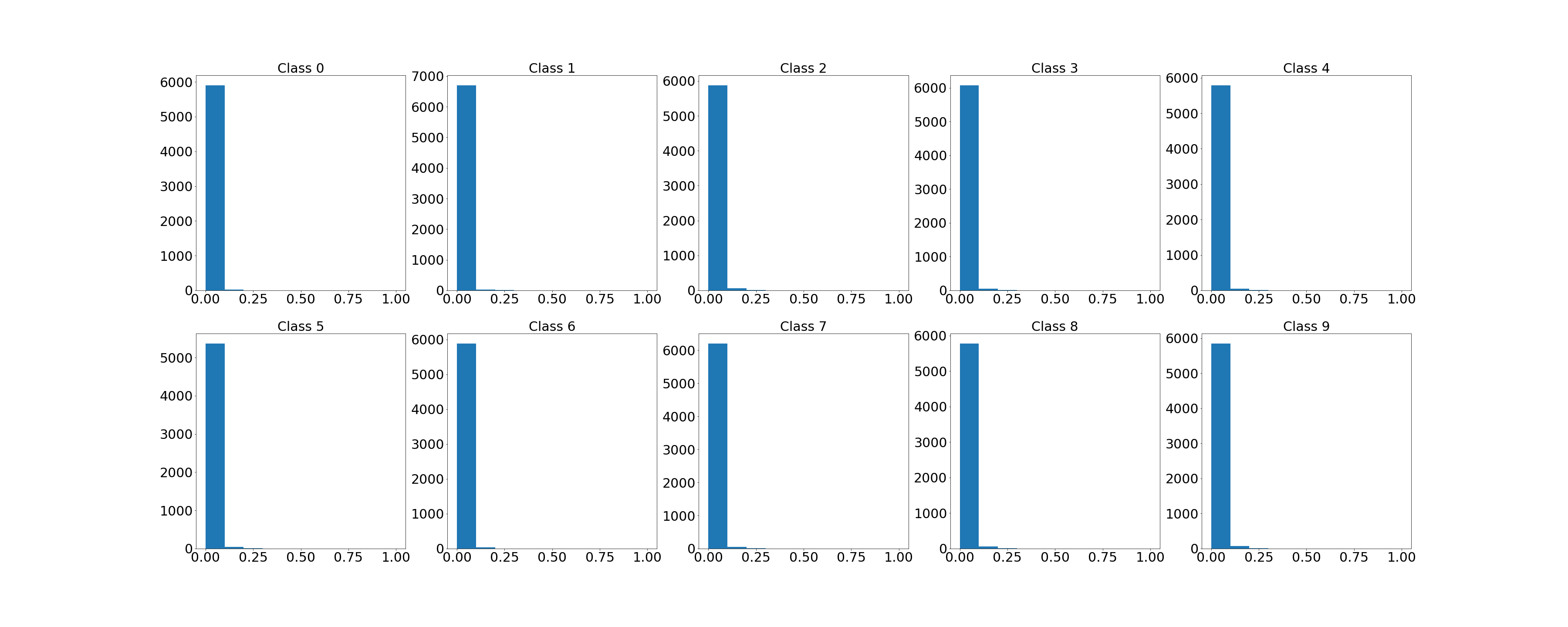}
    \caption{MNIST: Classwise sample significance extracted during teacher training with self regulation at $\alpha = 0.02$. x axis denotes the sample significance $\hat{v}$ and y axis denotes the frequency.}
    %\label{fig:my_label}
\end{figure*}

\begin{figure*}[]
    \centering
    \includegraphics[width=0.8\linewidth]{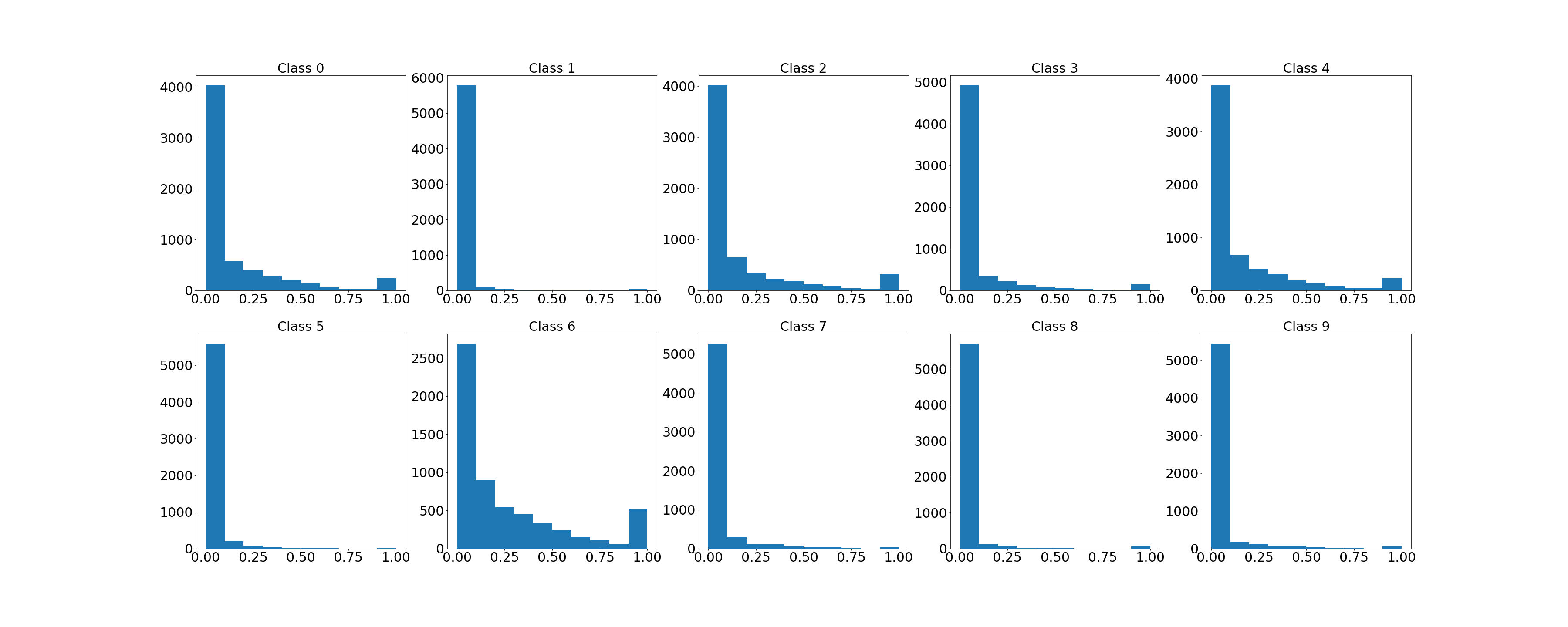}
    \caption{FMNIST: Classwise sample significance extracted during teacher training with self regulation at $\alpha = 0.04$. x axis denotes the sample significance $\hat{v}$ and y axis denotes the frequency.}
    %\label{fig:my_label}
\end{figure*}

\begin{figure*}[]
    \centering
    \includegraphics[width=0.8\linewidth]{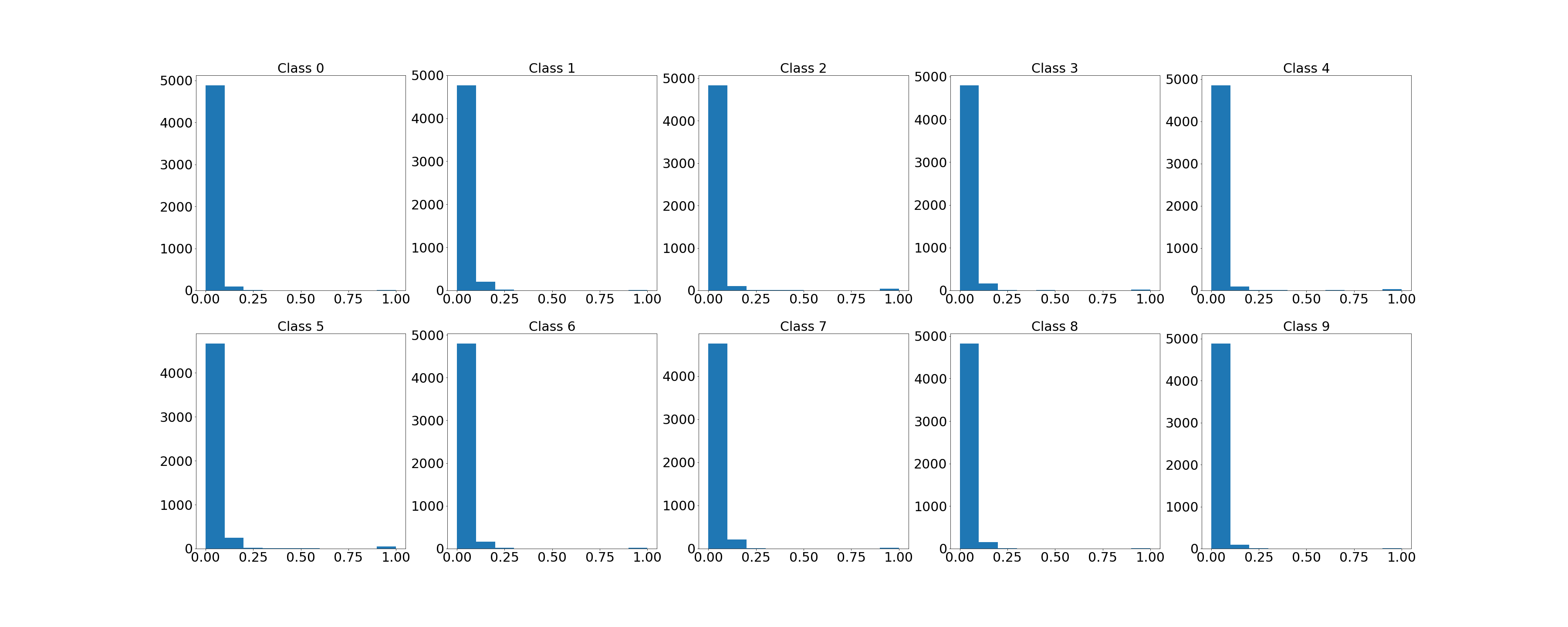}
    \caption{CIFAR10: Classwise sample significance extracted during teacher training with self regulation at $\alpha = 0.04$. x axis denotes the sample significance $\hat{v}$ and y axis denotes the frequency.}
    %\label{fig:my_label}
\end{figure*}

\subsection{Evaluation of Sample Efficiency}
To establish the data efficiency of the proposed self-regulated training method, the sample significance $\hat{v}$ extracted during teacher training is visualized as a histogram for each class. Figures 3-5 show these. These numbers are used as weights in the 'significance based distillation' and the 'hybrid distillation' processes.

Most samples are insignificant towards learning as indicated by the large frequency bars in the 0.0-0.25 bin on each plot. This is because the teacher model learns fast on these samples, so they participate less often in training. For the Fashion-MNIST dataset, classes 0, 2, 4, and 6 have a similar shape of the sample significance histograms. These class indices correspond to T-Shirt, Coat, Pullover, and Shirt classes. Since these objects have similar appearances, the model needs to see them more often to be able to classify them properly.

Mathematically, the sample efficiency $\zeta$ is defined as:

\begin{equation}
    \zeta = \frac{\sum_{i=1}^t \mathbf{v}[i]}{N|\mathbb{D}|} = \frac{\sum_{i=1}^t \mathbf{v}[i]}{Nt}
\end{equation}

where $\mathbf{v}$ is the array of sample participations used in Algorithms 1 and 2.

Since the 'significance based distillation' process is similar to the conventional distillation process \cite{normal}, its data efficiency is not evaluated. The total sample participation across all epochs for the 'regulated distillation' and the 'hybrid distillation' processes are reported. It is also reported as a percentage of all samples available for distillation across all epochs. This helps to compare the data efficiency of the proposed methods relative to the conventional distillation process \cite{normal}. The results are tabulated in Table 6. The first number denotes the total sample participation in the distillation process across all epochs. The second number denotes all the samples available for distillation across all epochs. The number in the parentheses is the first number expressed as a percentage of the second number. For example, the total sample participation across all epochs is 85528 for the MNIST dataset in the 'regulated distillation' process. However, distillation is performed for 200 epochs and 60000 samples are available for it in every epoch, making a total of 12000000 samples. This would be the total sample participation across all epochs for a normal distillation process \cite{normal}. So, 85528 is reported as a percentage of 12000000. This is the sample efficiency $\zeta$. 

Sample participation is relatively higher for the hybrid method. This is because the student model learns through self-regulation while using the sample significance data ($\hat{v}$, used as weights) obtained during teacher training. 

\begin{table}[]
\caption{Sample efficiency $\zeta$ of proposed distillations.}
\centering
\begin{tabular}{|l|l|l|}
\hline
\textbf{Dataset} & \textbf{Regulated Distillation}                                              & \textbf{\begin{tabular}[c]{@{}l@{}} Hybrid Distillation\end{tabular}} \\ \hline
\textbf{MNIST}   & \begin{tabular}[c]{@{}l@{}}85528/12000000 \\ ($\sim$0.713\%)\end{tabular}    & \begin{tabular}[c]{@{}l@{}}250245/12000000 \\ ($\sim$2.085\%)\end{tabular}            \\
\textbf{FMNIST}  & \begin{tabular}[c]{@{}l@{}}1385966/12000000 \\ ($\sim$11.549\%)\end{tabular} & \begin{tabular}[c]{@{}l@{}}2096926/12000000 \\ ($\sim$17.474\%)\end{tabular}          \\
\textbf{CIFAR10} & \begin{tabular}[c]{@{}l@{}}4553084/50000000 \\ ($\sim$9.106\%)\end{tabular}  & \begin{tabular}[c]{@{}l@{}}4605764/50000000 \\ ($\sim$9.211\%)\end{tabular}           \\ \hline

\end{tabular}
\label{table:Table 3}
\end{table}

In addition to being data-efficient, the proposed methods perform comparable to other state-of-the-art data-free methods (as shown in Tables 3-5) for distillation. The original training data is used as the transfer set and the sample participation shows that the proposed methods use much less data ($<$ 20\%) for distillation and training in general while achieving similar or better performance compared to other state-of-the-art methods. 

\section{Conclusions}
Data efficiency is a significant drawback of the existing distillation methods. Data efficiency improves by incorporating self-regulation in the training process. With self-regulated training, models can achieve similar generalization levels as if they were trained conventionally, with fewer samples. This finding shows that all the samples present in the training set are not equally important towards learning. This modification also improves the distillation performance in general as student models reach similar or better levels of generalization as other state-of-the-art methods, with fewer samples. The significance values obtained from the teacher model's self-regulated training help the students to generalize better. Regulated and hybrid variants of distillation are better suited to the knowledge transfer process in more realistic scenarios as the student has the freedom to learn on its own through self-regulation. Experiments on benchmark datasets establish the data efficacy of the proposed distillation methods (these use $<$ 20\% of the training data during distillation) and their competitive performance with other state-of-the-art distillation methods. 

The proposed methods do not indicate the minimum number of samples sufficient for transferring a certain level of generalization ability from the teacher to the student. In this direction, more studies will be conducted to determine how many samples from the training set are sufficient to represent the knowledge of a network to a given extent. Furthermore, the extension of the self-regulation approach to generative models will also be explored so that significant samples can be constructed from a given pre-trained model. This will make the proposed approaches data-free by constructing significant samples in a zero-shot fashion. The sensitivity of the significance values to the order in which the samples are presented for training will also be investigated. Finally, instead of employing a heuristic function to implement self-regulation, another network can be used to identify the significant samples along with the training process in an end-to-end manner.

\section*{Acknowledgment}
The authors would like to acknowledge WIRIN (WIPRO IISc Research Initiative) for the financial support.

\bibliographystyle{IEEEtran}
\bibliography{refs}

\begin{thebibliography}{10}
\providecommand{\url}[1]{#1}
\csname url@samestyle\endcsname
\providecommand{\newblock}{\relax}
\providecommand{\bibinfo}[2]{#2}
\providecommand{\BIBentrySTDinterwordspacing}{\spaceskip=0pt\relax}
\providecommand{\BIBentryALTinterwordstretchfactor}{4}
\providecommand{\BIBentryALTinterwordspacing}{\spaceskip=\fontdimen2\font plus
\BIBentryALTinterwordstretchfactor\fontdimen3\font minus
  \fontdimen4\font\relax}
\providecommand{\BIBforeignlanguage}[2]{{%
\expandafter\ifx\csname l@#1\endcsname\relax
\typeout{** WARNING: IEEEtran.bst: No hyphenation pattern has been}%
\typeout{** loaded for the language `#1'. Using the pattern for}%
\typeout{** the default language instead.}%
\else
\language=\csname l@#1\endcsname
\fi
#2}}
\providecommand{\BIBdecl}{\relax}
\BIBdecl

\bibitem{NIPS2012_c399862d}
A.~Krizhevsky, I.~Sutskever, and G.~E. Hinton, ``{ImageNet Classification with
  Deep Convolutional Neural Networks},'' in \emph{Advances in Neural
  Information Processing Systems}, vol.~25, 2012, pp. 1097--1105.

\bibitem{obj}
S.~Ren, K.~He, R.~Girshick, and J.~Sun, ``{Faster R-CNN: Towards Real-Time
  Object Detection with Region Proposal Networks},'' \emph{IEEE Transactions on
  Pattern Analysis and Machine Intelligence}, 2017.

\bibitem{Han2015}
S.~Han, J.~Pool, J.~Tran, and W.~J. Dally, ``{Learning both weights and
  connections for efficient neural networks},'' in \emph{Advances in Neural
  Information Processing Systems}, 2015.

\bibitem{Han2016}
S.~Han, H.~Mao, and W.~J. Dally, ``{Deep compression: Compressing deep neural
  networks with pruning, trained quantization and Huffman coding},'' in
  \emph{4th International Conference on Learning Representations, ICLR 2016 -
  Conference Track Proceedings}, 2016.

\bibitem{Sandryhaila2013}
A.~Sandryhaila and J.~M. Moura, ``{Discrete signal processing on graphs},''
  \emph{IEEE Transactions on Signal Processing}, 2013.

\bibitem{Denton2014}
E.~Denton, W.~Zaremba, J.~Bruna, Y.~LeCun, and R.~Fergus, ``{Exploiting linear
  structure within convolutional networks for efficient evaluation},'' in
  \emph{Advances in Neural Information Processing Systems}, 2014.

\bibitem{Wang2016}
Y.~Wang, C.~Xu, S.~You, D.~Tao, and C.~Xu, ``{CNNpack: Packing convolutional
  neural networks in the frequency domain},'' in \emph{Advances in Neural
  Information Processing Systems}, 2016.

\bibitem{normal}
G.~Hinton, O.~Vinyals, and J.~Dean, ``{Distilling the Knowledge in a Neural
  Network},'' in \emph{NIPS Deep Learning and Representation Learning
  Workshop}, 2015.

\bibitem{ZSKD}
G.~K. Nayak, K.~R. Mopuri, V.~Shaj, R.~{Venkatesh Babu}, and A.~Chakraborty,
  ``{Zero-shot knowledge distillation in deep networks},'' \emph{36th
  International Conference on Machine Learning, ICML 2019}, vol. 2019-June, pp.
  8317--8325, 2019.

\bibitem{FSKD}
A.~Kimura, Z.~Ghahramani, K.~Takeuchi, T.~Iwata, and N.~Ueda, ``{Few-shot
  learning of neural networks from scratch by pseudo example optimization},''
  \emph{British Machine Vision Conference 2018, BMVC 2018}, pp. 1--12, 2019.

\bibitem{DAFL}
H.~Chen, Y.~Wang, C.~Xu, Z.~Yang, C.~Liu, B.~Shi, C.~Xu, C.~Xu, and Q.~Tian,
  ``{Data-free learning of student networks},'' \emph{Proceedings of the IEEE
  International Conference on Computer Vision}, vol. 2019-Octob, pp.
  3513--3521, 2019.

\bibitem{brainDamage}
Y.~LeCun, J.~Denker, and S.~Solla, ``{Optimal Brain Damage},'' in
  \emph{Advances in Neural Information Processing Systems}, vol.~2, 1990.

\bibitem{Romero2015}
A.~Romero, N.~Ballas, S.~E. Kahou, A.~Chassang, C.~Gatta, and Y.~Bengio,
  ``{FitNets: Hints for thin deep nets},'' \emph{3rd International Conference
  on Learning Representations, ICLR 2015 - Conference Track Proceedings}, pp.
  1--13, 2015.

\bibitem{Heo2019}
B.~Heo, M.~Lee, S.~Yun, and J.~Y. Choi, ``{Knowledge transfer via distillation
  of activation boundaries formed by hidden neurons},'' \emph{33rd AAAI
  Conference on Artificial Intelligence, AAAI 2019}, pp. 3779--3787, 2019.

\bibitem{Zagoruyko2017}
S.~Zagoruyko and N.~Komodakis, ``{Paying more attention to attention: Improving
  the performance of convolutional neural networks via attention transfer},''
  \emph{5th International Conference on Learning Representations, ICLR 2017 -
  Conference Track Proceedings}, pp. 1--13, 2017.

\bibitem{meta}
R.~G. Lopes, S.~Fenu, and T.~Starner, ``{Data-Free Knowledge Distillation for
  Deep Neural Networks},'' \emph{LLD Workshop at Neural Information Processing
  Systems (NIPS)}, 2017.

\bibitem{meta1}
G.~{Sateesh Babu} and S.~Suresh, ``{Meta-cognitive Neural Network for
  classification problems in a sequential learning framework},''
  \emph{Neurocomputing}, 2012.

\bibitem{meta2}
R.~Savitha, S.~Suresh, and N.~Sundararajan, ``{Metacognitive learning in a
  fully complex-valued radial basis function neural network},'' \emph{Neural
  Computation}, 2012.

\bibitem{meta3}
G.~S. Babu and S.~Suresh, ``{Meta-cognitive RBF Network and its Projection
  Based Learning algorithm for classification problems},'' \emph{Applied Soft
  Computing Journal}, 2013.

\bibitem{org}
C.~Bucilǎ, R.~Caruana, and A.~Niculescu-Mizil, ``{Model compression},'' in
  \emph{Proceedings of the ACM SIGKDD International Conference on Knowledge
  Discovery and Data Mining}, 2006.

\bibitem{deepFace}
Y.~{Feng}, H.~{Wang}, H.~R. {Hu}, L.~{Yu}, W.~{Wang}, and S.~{Wang}, ``Triplet
  distillation for deep face recognition,'' in \emph{2020 IEEE International
  Conference on Image Processing (ICIP)}, 2020, pp. 808--812.

\bibitem{CMHash}
H.~{Hu}, L.~{Xie}, R.~{Hong}, and Q.~{Tian}, ``Creating something from nothing:
  Unsupervised knowledge distillation for cross-modal hashing,'' in \emph{2020
  IEEE/CVF Conference on Computer Vision and Pattern Recognition (CVPR)}, 2020,
  pp. 3120--3129.

\bibitem{collaborative}
Q.~{Guo}, X.~{Wang}, Y.~{Wu}, Z.~{Yu}, D.~{Liang}, X.~{Hu}, and P.~{Luo},
  ``Online knowledge distillation via collaborative learning,'' in \emph{2020
  IEEE/CVF Conference on Computer Vision and Pattern Recognition (CVPR)}, 2020,
  pp. 11\,017--11\,026.

\bibitem{review}
\BIBentryALTinterwordspacing
J.~Gou, B.~Yu, S.~J. Maybank, and D.~Tao, ``{Knowledge Distillation: A
  Survey},'' 2020. [Online]. Available: \url{http://arxiv.org/abs/2006.05525}
\BIBentrySTDinterwordspacing

\bibitem{self-kd1}
T.~Furlanello, Z.~C. Lipton, M.~Tschannen, L.~Itti, and A.~Anandkumar,
  ``{Born-Again Neural Networks},'' \emph{35th International Conference on
  Machine Learning, ICML 2018}, vol.~4, pp. 2615--2624, 2018.

\bibitem{adam}
D.~P. Kingma and J.~L. Ba, ``{Adam: A method for stochastic optimization},''
  \emph{3rd International Conference on Learning Representations, ICLR 2015 -
  Conference Track Proceedings}, pp. 1--15, 2015.

\bibitem{LeCun1998}
Y.~LeCun, L.~Bottou, Y.~Bengio, and P.~Haffner, ``{Gradient-based learning
  applied to document recognition},'' \emph{Proceedings of the IEEE}, 1998.

\end{thebibliography}
% \nocite{*}
% \bibitem{b1} G. Eason, B. Noble, and I. N. Sneddon, ``On certain integrals of Lipschitz-Hankel type involving products of Bessel functions,'' Phil. Trans. Roy. Soc. London, vol. A247, pp. 529--551, April 1955.
% \bibitem{b2} J. Clerk Maxwell, A Treatise on Electricity and Magnetism, 3rd ed., vol. 2. Oxford: Clarendon, 1892, pp.68--73.
% \bibitem{b3} I. S. Jacobs and C. P. Bean, ``Fine particles, thin films and exchange anisotropy,'' in Magnetism, vol. III, G. T. Rado and H. Suhl, Eds. New York: Academic, 1963, pp. 271--350.
% \bibitem{b4} K. Elissa, ``Title of paper if known,'' unpublished.
% \bibitem{b5} R. Nicole, ``Title of paper with only first word capitalized,'' J. Name Stand. Abbrev., in press.
% \bibitem{b6} Y. Yorozu, M. Hirano, K. Oka, and Y. Tagawa, ``Electron spectroscopy studies on magneto-optical media and plastic substrate interface,'' IEEE Transl. J. Magn. Japan, vol. 2, pp. 740--741, August 1987 [Digests 9th Annual Conf. Magnetics Japan, p. 301, 1982].
% \bibitem{b7} M. Young, The Technical Writer's Handbook. Mill Valley, CA: University Science, 1989.
% \end{thebibliography}
% \vspace{12pt}
% \color{red}
% IEEE conference templates contain guidance text for composing and formatting conference papers. Please ensure that all template text is removed from your conference paper prior to submission to the conference. Failure to remove the template text from your paper may result in your paper not being published.

\end{document}